\documentclass[10pt,twocolumn,letterpaper]{article}

\usepackage{cvpr}
\usepackage{times}
\usepackage{epsfig}
\usepackage{graphicx}
\usepackage{amsmath}
\usepackage{amssymb}
\usepackage{multirow}
\usepackage{booktabs}

\cvprfinalcopy 

\def\cvprPaperID{****} 

\begin{document}

\title{Differentiable Neural Architecture Transformation\\for Reproducible Architecture Improvement}

\author{Do-Guk Kim\thanks{Equal contribution}$~~^\dagger$,~~Heung-Chang Lee$^\ast$\thanks{Corresponding authors}\\
BigData \& AI Lab, Hana Institute of Technology, Hana TI\\
Seoul, Korea\\
{\tt\small logue311@gmail.com, leehc.com@gmail.com}}

\maketitle


\begin{abstract}
Recently, Neural Architecture Search (NAS) methods are introduced and show impressive performance on many benchmarks.
Among those NAS studies, Neural Architecture Transformer (NAT) aims to improve the given neural architecture to have better performance while maintaining computational costs.
However, NAT has limitations about a lack of reproducibility.
In this paper, we propose differentiable neural architecture transformation that is reproducible and efficient.
The proposed method shows stable performance on various architectures.
Extensive reproducibility experiments on two datasets, i.e., CIFAR-10 and Tiny Imagenet, present that the proposed method definitely outperforms NAT and be applicable to other models and datasets.
\end{abstract}

\section{Introduction}
\label{Intro}
Neural architectures designed by Neural Architecture Search (NAS) algorithms achieved state-of-the-art performances on many benchmark datasets.
Despite the great performance, NAS methods are hard to use because of their prohibitively high computational costs.
Therefore, many recent works focused on reducing the computational costs of NAS while maintaining the advantage of NAS approaches.

Neural Architecture Transformer (NAT)~\cite{guo2019nat} is the one of such kind of works.
The authors introduced the architecture transformation concept that requires less computational costs than traditional NAS methods.
Neural architecture transformation means that optimizing the performance of the network by modifying the operations while maintaining or reducing the computational costs.
In this work, the authors transform the original operation of a given neural architecture into only identity operation or none operation to achieve better performance or less computational costs.

Although they showed the possibility of the NAT to be used for network performance improvement, NAT has several limitations.
First, the reproducibility of the algorithm is not verified since the authors reported only one result for each model.
Second, the architecture transformation stage and network train stage is totally separated.
It requires not only a lot of computational resources but also an additional human effort to get a transformed architecture and train a neural network.
Third, it can only transform the neural networks with identical cell architectures.
Recent NAS works focus on searching macroblock based architectures that have various cell architectures.
However, those architectures cannot be transformed by NAT.

In this paper, we propose differentiable neural architecture transformation method that overcomes those limitations.
We claim the following contributions: 
\begin{itemize}
    \item We carried out extensive reproducibility experiments, and the results demonstrate the high reproducibility of the proposed method.
    \item We propose consecutive architecture transformation and network learning. The proposed method automatically transforms the architecture, trains the network, and outputs the trained networks.
    \item The proposed method can transform not only identical cell architectures but also full network architectures with various cell architectures like ProxylessNAS~\cite{cai2018proxylessnas}.
\end{itemize}

\begin{figure*}[t!]
\begin{center}
\centerline{\includegraphics[width=0.9\linewidth]{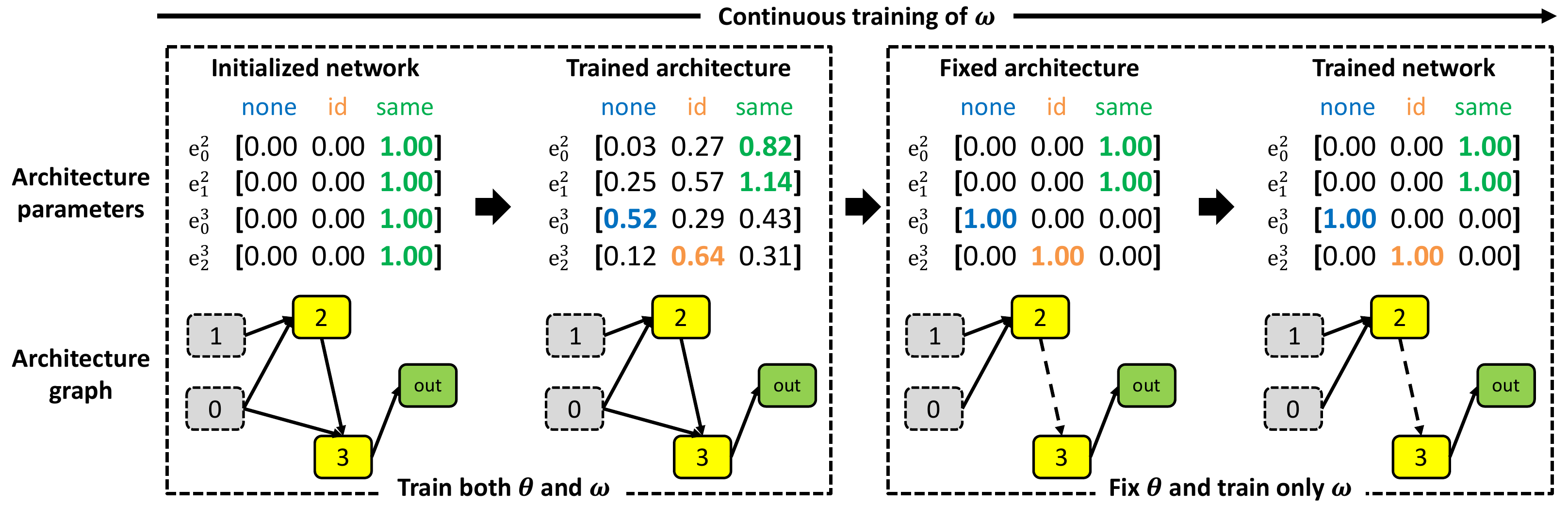}}
\caption{An example of the network architecture reforging by the proposed method. Until the training of the architecture parameters $\theta$ is finished, both $\omega$ and $\theta$ are trained. After the architecture train step, $\theta$ is fixed and only $\omega$ is trained until we get the final trained network.}
\vspace{-0.5cm}
\label{overall}
\end{center}
\end{figure*}

\section{Related Work}
\label{Related}
Since the NAS is introduced by~\cite{zoph2016neural}, many methods have been proposed to search effective neural architecture for a given dataset.
ENAS~\cite{pham2018efficient} presented shared weights that dramatically reduced the computational complexity of the NAS.
DARTS~\cite{liu2018darts} and NAO~\cite{luo2018neural} introduced gradient-based NAS schemes that search neural architecture by the gradient of architectural parameters and does not need the additional controller.

Recently, NAT~\cite{guo2019nat} proposed the architecture transformation concept that optimizes given neural architecture.
Unlike traditional NAS methods that search network architecture by selecting various operations, NAT only transforms the original operations into none or identity operations.
Although the authors showed impressive results in the paper, there are several drawbacks of the NAT we claimed in the Section~\ref{Intro}.

\section{Methodology}
\label{Method}

The proposed method improves the performance of the given neural architecture by using gradient-based optimization.
After the entire learning process, we directly get the trained network, and there is no need to train a new network from scratch.
There are two consecutive stages in the proposed method: architecture train stage and network train stage.
In the architecture train stage, both the architecture parameters and network parameters are trained.
After the architecture train stage, only network parameters are trained.
The overall process of the proposed method is shown in Figure~\ref{overall}.

\subsection{Differentiable architecture parameters}
Unlike NAT algorithm used Graph Convolutional Network (GCN) and reinforcement learning, we use differentiable architecture parameters and gradient-based learning.
The architecture parameters $\theta$ is defined in the network architecture graph.
Each edge in the network architecture graph contains original operation, identity operation, and none operation.
Computation of each edge is carried out based on the architecture parameters:
\begin{equation}
\label{arch_params}
    {o}_{e}(x)={\theta}_{e, none}\cdot Z + {\theta}_{e, id}\cdot x + {\theta}_{e, same}\cdot o(x),
\end{equation}
where $x$ means input, ${o}_{e}(x)$ means output of the edge, $Z$ means zero tensor, $o(x)$ means original operation of the edge, ${\theta}_{e, none}$ means the weight of the none operation of the edge, ${\theta}_{e, id}$ means the weight of the identity operation of the edge, and ${\theta}_{e, same}$ means the weight of the original operation of the edge.
We set initial ${\theta}_{none}$ and ${\theta}_{id}$ as zero, and ${\theta}_{same}$ as one.
Therefore, the initialized network works the same as the original architecture.
An example of architecture parameters is presented in Figure~\ref{overall}.
There are four edges in the cell architecture (except edge to output node), and three operations for each edge.
Therefore, the size of the parameters is $4\times 3$.
The proposed architecture parameters can be used for improving the full network architecture rather than the cell architecture.
Assume that there are eight cells in the entire network, and each cell has four computational edges, then the size of the architecture parameters for the full network becomes $32 \times 3$.

\subsection{Architecture train stage}
After the network is initialized, the architecture train stage begins to improve the given architecture.
In this stage, both the network weight parameters $\omega$ and the architecture parameters $\theta$ are trained alternately.
For every input mini-batches, $\omega$ is trained first, and $\theta$ is trained after the update of $\omega$.
Note that the proposed method doesn't require any separated dataset for architecture optimization, and the network can utilize a full dataset to train its weights $\omega$.
The architecture train stage is carried out for the pre-defined epochs.

When the architecture train stage is finished, architecture is transformed based on the trained $\theta$.
For each edge, the operation that has the highest weight in $\theta$ is selected to construct the final architecture.
In the example of Figure~\ref{overall}, two edges maintain the original operations, one edge changed its operation into identity, and one edge is removed because none operation is selected.

\subsection{Network train stage}
The architecture trained at the previous stage is fixed, and only $\omega$ of the network is trained in this stage.
This stage is the same as the traditional neural network training process, and it is continuously carried out after the architecture train stage.
At the end of this stage, we can get the trained network and use it to infer unseen input data or test the performance of the model.

\begin{table}[t]
\caption{Comparison of Average Accuracy, Standard Deviation and Total Cost between original, NAT and Ours on CIFAR-10.}
\label{tab_cifar10_result}
\vskip 0.05in
\begin{center}
\scalebox{0.8}{
\begin{tabular}{clccccc}
\toprule
& \multicolumn{1}{c}{}& \multicolumn{1}{c}{Avg Acc} & \multicolumn{1}{c}{Std} & \multicolumn{1}{c}{Total Cost}\\
Model & \multicolumn{1}{l}{Method} & \multicolumn{1}{c}{(\%)} & \multicolumn{1}{c}{(\%)}& \multicolumn{1}{c}{(GPU hours)}\\

\midrule
\multirow{4}{*}{Resnet20~\cite{he2016deep}} & Original & 91.66 & 0.16 & 8\\
& NAT ~\cite{guo2019nat} & 55.78 & 42.34 &  14\\
\cmidrule{2-5}
& Ours(Cell) & \textbf{93.29} & 0.11 & 11.2\\
& Ours(Full) & 93.12 & 0.12 & 8.7\\
\midrule
\multirow{4}{*}{\shortstack{Mobilenet \\ V2~\cite{sandler2018mobilenetv2}}} 
& Original & 93.91 & 0.12 & 20.5\\
& NAT ~\cite{guo2019nat} & 91.97 & 5.10 & 27.8 \\
\cmidrule{2-5}
& Ours(Cell) & \textbf{95.02} & 0.31 & 24.8\\
& Ours(Full) & 94.93 & 0.13 & 22.1\\
\midrule
\midrule
\multirow{4}{*}{DARTS~\cite{liu2018darts}} & Original & 96.75 & 0.11 & 38.3\\
& NAT ~\cite{guo2019nat} & 96.95 & 0.09 & 47 \\
\cmidrule{2-5}
& Ours(Cell) & \textbf{96.97} & 0.15 & 45\\
& Ours(Full) & 96.82 & 0.13 & 41.1\\
\midrule
\multirow{3}{*}{\shortstack{Proxyless \\ NAS~\cite{cai2018proxylessnas}}} & Original & 94.19 & 1.08 & 15.3\\
& NAT ~\cite{guo2019nat} & - & - & - \\
\cmidrule{2-5}
& Ours(Full) & \textbf{95.09} & 0.23 & 19.8\\
\bottomrule
\end{tabular}
}
\end{center}
\vskip -0.1in
\end{table}

\begin{table}[t]
\caption{Reproducibility of original, NAT and Ours with different random seeds on CIFAR-10.}
\label{tab_reproducibility}
\vskip 0.05in
\begin{center}
\scalebox{0.8}{
\begin{tabular}{clccccc}
\toprule
& \multicolumn{1}{c}{}& \multicolumn{5}{c}{Random Seed}\\
Model & \multicolumn{1}{l}{Method} & \multicolumn{1}{c}{(1)} & \multicolumn{1}{c}{(2)}& \multicolumn{1}{c}{(3)}& \multicolumn{1}{c}{(4)}& \multicolumn{1}{c}{(5)}\\

\midrule
\multirow{4}{*}{Resnet20~\cite{he2016deep}} & Original & 91.74 & 91.74 & 91.65 & 91.76 & 91.39\\
& NAT ~\cite{guo2019nat} & 10 & 75.14 & 91.05 & 10 & 92.68 \\
\cmidrule{2-7}
& Ours(Cell) & 93.21 & 93.34 & 93.4 & 93.15 & 93.37\\
& Ours(Full) & 93.07 & 93.26 & 93.22 & 93.06 & 92.97\\
\midrule
\multirow{4}{*}{\shortstack{Mobilenet \\ V2~\cite{sandler2018mobilenetv2}}} 
& Original & 93.9 & 94.04 & 93.95 & 93.95 & 93.72\\
& NAT ~\cite{guo2019nat} & 83.36 & 95.13 & 94.95 & 91.18 & 95.21 \\
\cmidrule{2-7}
& Ours(Cell) & 94.57 & 94.97 & 95.41 & 94.97 & 95.18\\
& Ours(Full) & 95.13 & 94.81 & 94.85 & 95 & 94.85\\
\bottomrule
\end{tabular}
}
\end{center}
\vskip -0.1in
\end{table}

\section{Experiments}
\label{Experiments}
We carried out extensive experiments to verify the performance and the reproducibility of comparison methods.
In the experiments, various models are trained on CIFAR-10 and Tiny Imagenet.

\subsection{Data and Experiment Setting}
CIFAR-10 dataset consists of 50,000 train images and 10,000 test images with ten classes.
The size of images is $32\times32$, and images have RGB color channels.
Tiny Imagenet dataset has 100,000 train images and 10,000 test images with 200 classes.
The input size of Tiny Imagenet dataset is $64\times64$, and all images are RGB color images.

We experimented with various models on CIFAR-10 and Tiny Imagenet dataset.
These models include ResNet20, MobileNet V2, DARTS, and ProxylessNAS.
Former two models are manually designed, and the latter two models are NAS models.
To compare the performance of NAT and our algorithm, we trained NAT controller on each dataset and then trained the transformed architecture inferred from the controller.
In the case of our algorithm, we test both cell-based transformation and full network transformation.
We used 0.025 learning rate, 600 epoch, and Stochastic Gradient Descent(SGD) optimizer as the same hyper-parameters to all models and methods.
Exceptionally, we applied 300 epoch to Mobilenet V2 and DARTS on Tiny Imagenet dataset, and utilized cut-out for NAS models such as DARTS and ProxylessNAS.

We tested five times with different random seeds for every experiment to get the right performance and verify the reproducibility of comparison algorithms.
Therefore, we report the average accuracy and standard deviation of each method and each model.
The total cost of NAT was calculated by adding GPU hours of the architecture transformation stage and network train stage, and the cost of our algorithms was computed by just measure the cost of the whole training process.

\begin{figure}[t]
\begin{center}
\includegraphics[width=1.0\linewidth]{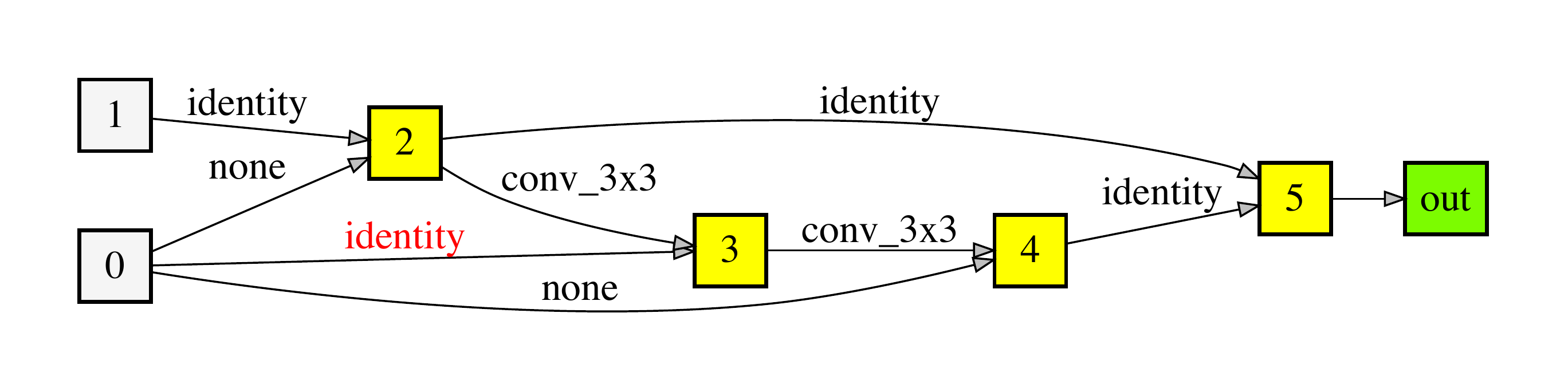}
\vspace{-0.2cm}
\centerline{\small (a) Ours}
\includegraphics[width=1.0\linewidth]{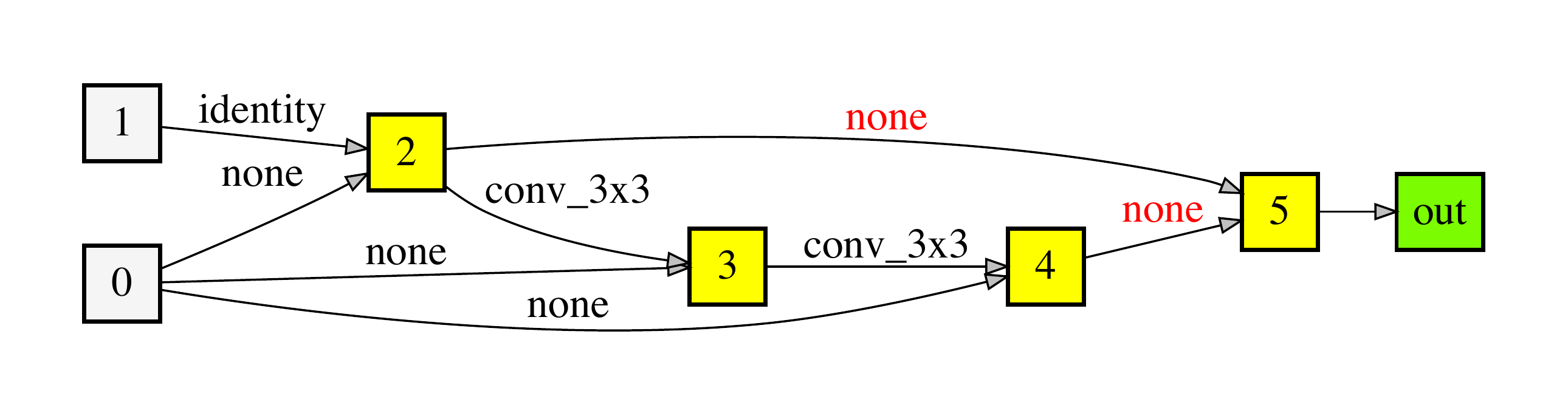}
\centerline{\small (b) NAT~\cite{guo2019nat}}
\vspace{-0.2cm}
\caption{Transformed ResNet20 cell architectures}
\vspace{-0.5cm}
\label{resnet}
\end{center}
\end{figure}

\begin{figure}[t]
\begin{center}
\includegraphics[width=1.0\linewidth]{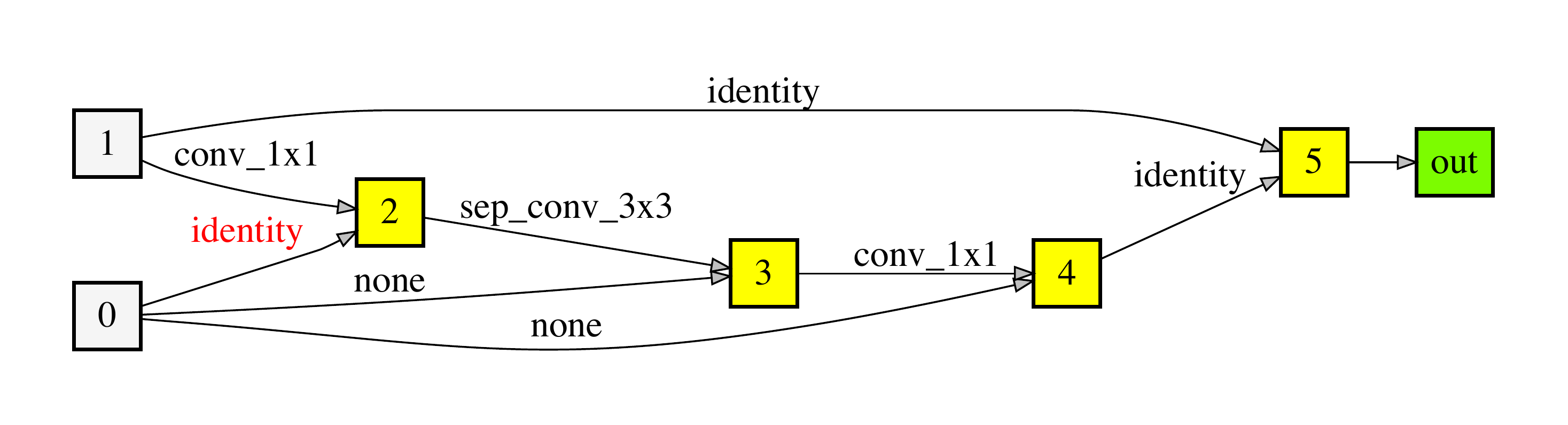}
\vspace{-0.2cm}
\centerline{\small (a) Ours}
\includegraphics[width=1.0\linewidth]{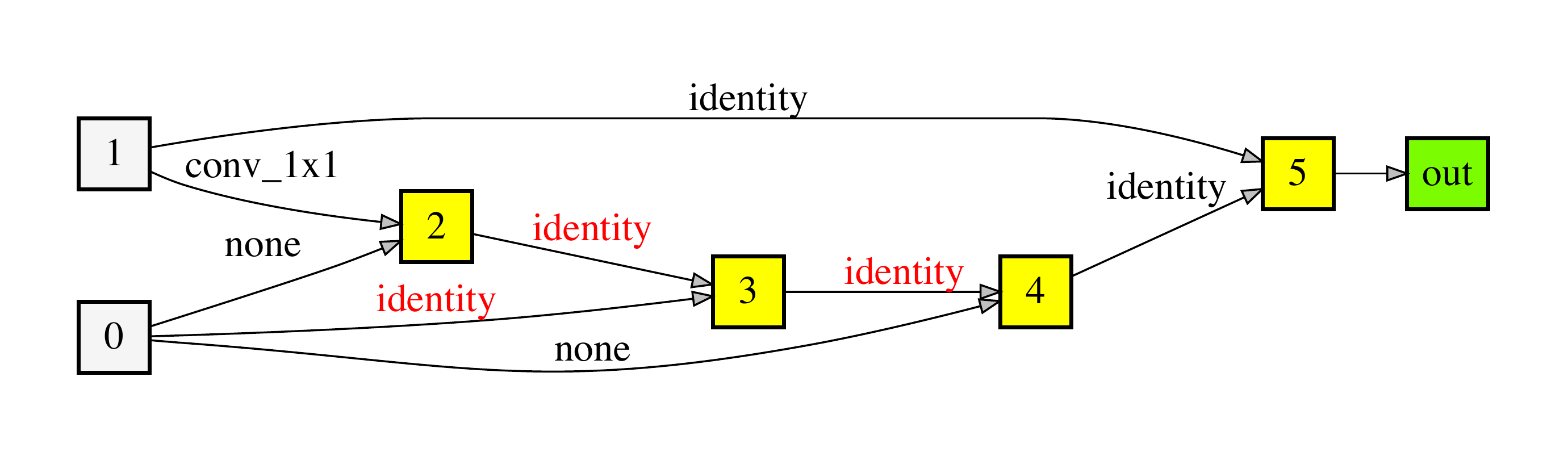}
\centerline{\small (b) NAT~\cite{guo2019nat}}
\vspace{-0.2cm}
\caption{Transformed MobileNetV2 cell architectures}
\vspace{-0.5cm}
\label{mobilenet}
\end{center}
\end{figure}

\begin{figure}[t]
\begin{center}
\includegraphics[width=0.7\linewidth]{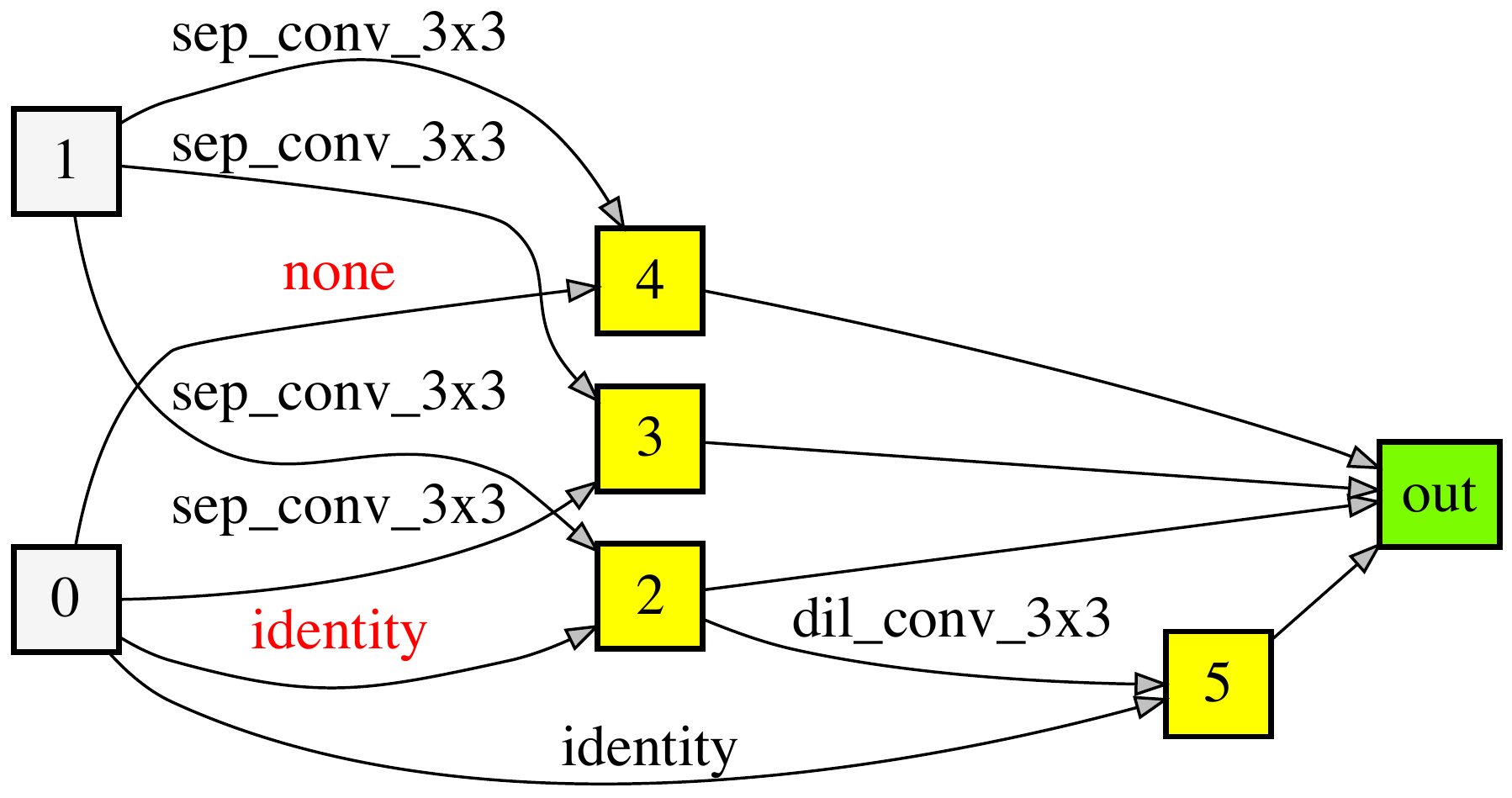}
\caption{DARTS normal cell transformed by our method. Reduction cell is the same as the original one.}
\label{darts}
\end{center}
\end{figure}

\subsection{Results and Discussion}
The results of Table~\ref{tab_cifar10_result} have average accuracy, standard deviation, and total cost by various methods with different models on CIFAR-10 dataset.
We trained and inferred five times to get average accuracy and standard deviation.
As shown in Table~\ref{tab_cifar10_result}, the results of NAT is unstable in the case of manually designed models.
The results of our algorithms have better average accuracy and standard deviation than original and NAT in all cases. 
Moreover, the total computational cost is lower than NAT. 
Note that NAT cannot transform the architecture of ProxylessNAS, since it has various cell architectures in the network.
However, the proposed method successfully improves the performance of ProxylessNAS architecture.

Table~\ref{tab_reproducibility} shows the reproducibility of various methods with different random seeds on CIFAR-10 dataset.
Only one result is presented for each model in NAT paper. Therefore we experimented five times to get the right performance.
In the case of Resnet20 experiments, the results of seed 1 and 4 of NAT are caused by transform identity edges into none operation.
Transformed Resnet20 architectures of seed 1 are presented in Figure~\ref{resnet}.
Changed edges are notated as red colors.
As shown in Figure~\ref{resnet}(b), NAT transformed all edges to node 5 into none operation.
Therefore, zero tensors are passed to the next layer.

Regarding the result of Mobilenet V2 experiments, the performance of NAT is degraded when it transforms convolution operation into identity operation.
Figure~\ref{mobilenet} shows the transformed Mobilenet V2 architectures of seed 1.
Additionally, we represent the transformed DARTS normal cell architecture of our algorithm in Figure~\ref{darts}.
There is no change in edges of reduction cell.

Table~\ref{tab_tiny_result} shows the results of various methods with different models on Tiny Imagenet dataset.
The results of this table describe that both our algorithms have better average accuracy, standard deviation, and faster GPU hours than the original method upon all models. 

\begin{table}[t]
\caption{Comparison Average Accuracy, Standard Deviation and Total Cost between original, NAT and Ours on Tiny Imagenet.}
\label{tab_tiny_result}
\vskip 0.05in
\begin{center}
\scalebox{0.8}{
\begin{tabular}{clccccc}
\toprule
& \multicolumn{1}{c}{}& \multicolumn{1}{c}{Avg Acc} & \multicolumn{1}{c}{Std} & \multicolumn{1}{c}{Total Cost}\\
Model & \multicolumn{1}{l}{Method} & \multicolumn{1}{c}{(\%)} & \multicolumn{1}{c}{(\%)}& \multicolumn{1}{c}{(GPU hours)}\\

\midrule
\multirow{3}{*}{Resnet20~\cite{he2016deep}} & Original & 50.72 & 0.41 & 16.1\\
\cmidrule{2-5}
& Ours(Cell) & \textbf{52.86} & 0.49 & 22\\
& Ours(Full) & 52.7118 & 0.23 & 17.5\\
\midrule
\multirow{3}{*}{\shortstack{Mobilenet \\ V2~\cite{sandler2018mobilenetv2}}} 
& Original & 51.57 & 0.76 & 20.3\\
\cmidrule{2-5}
& Ours(Cell) & \textbf{53.17} & 1.03 & 25.5\\
& Ours(Full) & 52.92 & 0.50 & 23.7\\
\midrule
\midrule
\multirow{3}{*}{DARTS~\cite{liu2018darts}} & Original & 59.25 & 0.44 & 39.2\\
\cmidrule{2-5}
& Ours(Cell) & 60.24 & 0.35 & 47\\
& Ours(Full) & \textbf{60.63} & 0.50 & 43.8\\
\bottomrule
\end{tabular}
}
\end{center}
\vskip -0.1in
\end{table}

\section{Conclusion}
\label{Conclusion}

We proposed a novel gradient-based neural architecture transformation algorithm that is reproducible and effective for architecture improvement.
Thanks to the differentiable architecture parameters, our algorithm can train both the architecture and the network at once.
The results of five times experiments of all methods demonstrate that the proposed algorithm has high reproducibility and stably improve the performance of various models on various datasets.

{\small
\bibliographystyle{ieee_fullname}
\bibliography{egpaper_final}
}

\end{document}